\documentclass[runningheads]{llncs}

\usepackage[FINAL,year=2024,ID=10629]{eccv}
\usepackage{eccvabbrv}

\usepackage{graphicx}
\usepackage{booktabs}
\usepackage{colortbl} % For colored rules
\usepackage{xcolor} % For coloring
\usepackage{lipsum} % for generating dummy text
\usepackage{amsmath} % assumes amsmath package installed
\usepackage{amssymb}  % assumes amsmath package installed

\usepackage[accsupp]{axessibility}  % Improves PDF readability for those with disabilities.

\usepackage{multirow}
\usepackage{lipsum} % for generating sample text
\usepackage{tikz}
\usepackage{atbegshi} % to hook into the shipout mechanism
\usepackage{svg}

\usepackage[pagebackref,breaklinks,colorlinks]{hyperref}
\usepackage{orcidlink}

\begin{document}

\title{Reframe Anything: LLM Agent for Open World Video Reframing} 

\titlerunning{Abbreviated paper title}

\author{Jiawang Cao\inst{1}\thanks{Equal contribution.} \and
Yongliang Wu\inst{1,2}\textsuperscript{$\star$} \and
Weiheng Chi\inst{1,3}\textsuperscript{$\star$} \and
Wenbo Zhu\inst{1}\textsuperscript{$\star$} \and
Ziyue Su\inst{1} \and
Jay Wu\inst{1}
}

\authorrunning{Jiawang~Cao et al.}

\institute{Opus AI Research \and
Southeast University \and
National University of Singapore \\
\email{\{gavin.cao, henry.chi, vito.zhu, lirian.su, jay.wu\}@opus.pro, yongliangwu@seu.edu.cn, weiheng\_chi@u.nus.edu}}

\maketitle

\begin{abstract}
The proliferation of mobile devices and social media has revolutionized content dissemination, with short-form video becoming increasingly prevalent. This shift has introduced the challenge of video reframing to fit various screen aspect ratios, a process that highlights the most compelling parts of a video. Traditionally, video reframing is a manual, time-consuming task requiring professional expertise, which incurs high production costs. A potential solution is to adopt some machine learning models, such as video salient object detection, to automate the process. However, these methods often lack generalizability due to their reliance on specific training data. The advent of powerful large language models (LLMs) open new avenues for AI capabilities. Building on this, we introduce \textbf{R}eframe \textbf{A}ny \textbf{V}ideo \textbf{A}gent (RAVA), a LLM-based agent that leverages visual foundation models and human instructions to restructure visual content for video reframing. RAVA operates in three stages: perception, where it interprets user instructions and video content; planning, where it determines aspect ratios and reframing strategies; and execution, where it invokes the editing tools to produce the final video. Our experiments validate the effectiveness of RAVA in video salient object detection and real-world reframing tasks, demonstrating its potential as a tool for AI-powered video editing.
\end{abstract}

\section{Introduction}
\label{sec:intro}
The short-form video has emerged as a novel and swiftly expanding mode of content dissemination under the rapid evolution of social media and handheld mobile devices~\cite{cochrane2014mobile}. Traditional video aspect ratios no longer cater to the convenience of social media platform viewing due to varying screen proportions. Consequently, the challenge of reconstructing original videos to accommodate different aspect ratios has become a burgeoning demand in the field of video editing. The process often involves identifying and focusing on the most captivating or crucial elements within the current frame. Additionally, from an artistic standpoint, it sometimes necessitates zooming in on a specific area of the scene. This technique is referred to as video reframing.

\begin{figure}[t]
    \centering
    \includegraphics[width=1\textwidth]{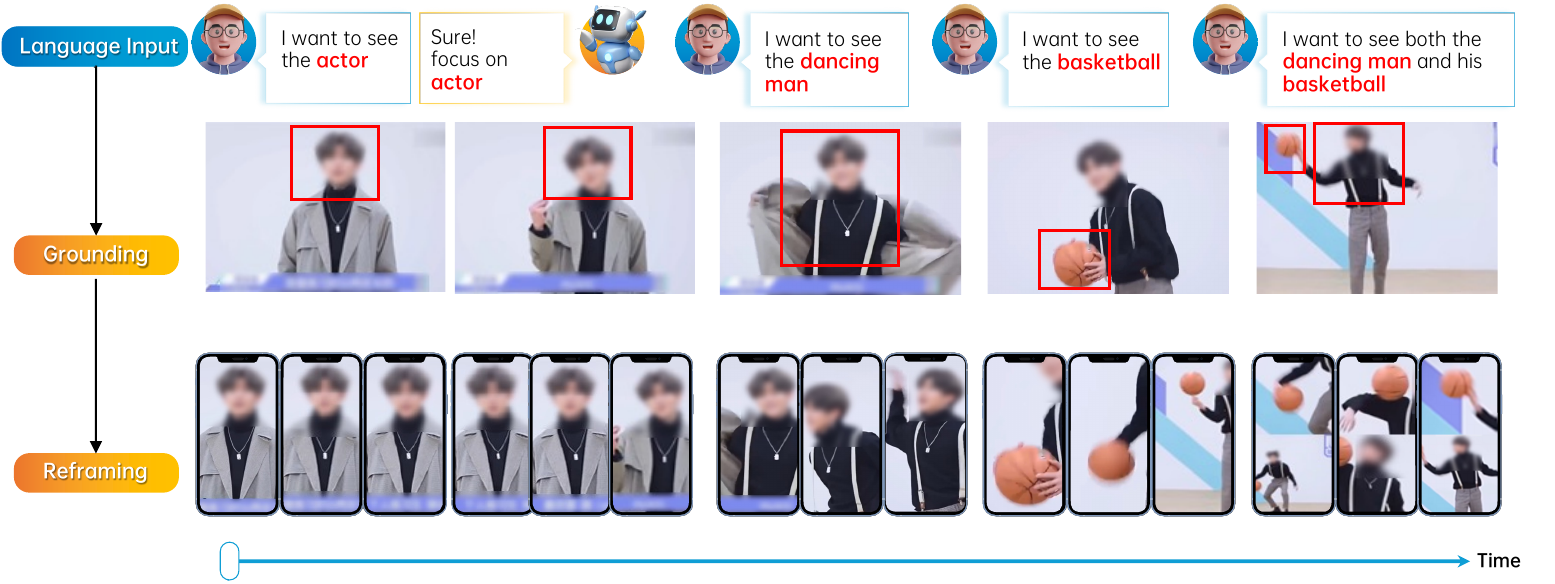}
    \caption{The overview of open world video reframing task. Even within the same video, different viewers may focus on different subjects of interest. Therefore, it is essential to implement video reframing based on user instructions to achieve specific objectives.}
    \label{fig:intro}
    \vspace{-5pt}
\end{figure}

Manual video reframing is a labor-intensive and time-consuming task that demands the expertise of professional editors. Creating high-quality videos typically involves skilled individuals, which in turn escalates the overall cost of production. Therefore, some machine learning researchers have started to investigate the possibility of automating video reframing. A practical strategy for this involves the application of video saliency detection~\cite{wang2019revisiting,jiang2018deepvs} which focuses on identifying the most salient, or attention-grabbing, region within a video. For example, Christel et al.~\cite{chamaret2008attention} utilize a bottom-up visual model to generate a saliency map of the current frame and edit the scene to focus on these key areas. The aforementioned methods, while partially useful, do not always ensure that the parts they extract are complete, which can hinder their use in real-world scenarios. In an attempt to address this limitation, research in video salient object detection make significant strides. This line of work focuses on segmenting the most visually striking objects in a video frame. By doing so, it facilitates a more accurate determination of the full scope of the segmented area. However, the effectiveness of these models is often compromised by their dependence on specific domains of training data. This dependency can limit their generalizability and negatively impact their performance and interpretability in diverse applications. Furthermore, taking into account that different viewers have varying interests in specific parts of a video and that users have diverse requirements for editing, as illustrated in Figure~\ref{fig:intro}, it is necessary to design a framework capable of flexibly performing video reframing according to user instructions.

Recently, the development of powerful large language models (LLMs), such as ChatGPT~\cite{openai2021chatgpt} and GPT-4~\cite{openai2023gpt4}, further propel the advancement of artificial intelligence. These models demonstrate a formidable capability for understanding and generating human language, including the ability to perceive and comprehend visual content through textual descriptions of images. For instance, they can convey the coordinates of objects within an image, grasping the essence of the visual scene, without the need for a direct vision encoder~\cite{liu2024visual}. The research community is swiftly moving towards utilizing LLMs as agents that can perform complex cognitive functions, which include perception, planning, and action execution. Pioneering developments in this area include systems like TaskMatrix~\cite{liang2023taskmatrix}, AutoGPT~\cite{yang2023auto}, and MetaGPT~\cite{hong2023metagpt}. Adding to this innovation, the advent of multimodal LLMs such as GPT-4V~\cite{openai2023gpt4v} usher in a new era where LLMs can directly sense visual content. Consequently, some works such as AppAgent~\cite{yang2023appagent} and MobileAgent~\cite{wang2024mobile} take this further by creating agents that can navigate and control any smartphone application within a mobile operating system. 

Inspired by these explorations into LLMs as agents, in this paper, we introduce \textbf{R}eframe \textbf{A}ny \textbf{V}ideo \textbf{A}gent (RAVA), a LLM-based agent designed to execute video reframing tasks flexibly based on human instructions. The overall framework can be divided into three main stages: perception, planning, and execution. In the perception phase, RAVA employs language learning to interpret user directives and video understanding to dissect scenes, identify pivotal objects, and generate textual scene descriptions. This grasp of content and context informs the subsequent planning stage, where the agent meticulously determines aspect ratios, prioritizes object importance, configures dynamic layouts, and devises a visual effect strategy that aligns with the narrative and user preferences. Finally, in the execution phase, RAVA translates the intricate plan into action, orchestrating the reframing with precision, applying effects, and arranging content according to a structured execution blueprint, all while allowing for feedback loops to refine the output. This comprehensive, three-stage process executed by RAVA ensures that videos are not only adapted to new formats but also resonate with the intended audience, amplifying the impact of content on various platforms.

Furthermore, to validate the effectiveness of RAVA, we embark on experiments from two dimensions. First, we apply it to the classic computer vision task of video salient object detection to verify its ability to accurately execute human instructions and its capability for scene comprehension. Second, we employ it in real-world video reframing tasks to determine its proficiency in completing this practically valuable task. Both quantitative results and user studies highlight the advantages of RAVA.

Our main contributions can be summarized as follows:
\begin{itemize}
\item We introduce RAVA, a LLM-based agent that is adept at performing video reframing tasks in accordance with human directives.
\item Through a carefully crafted perception, planning, and execution framework, RAVA is able to effectively utilize the power of existing foundational models to carry out human instructions accurately.
\item By conducting thorough experiments on video salient object detection and real-world video reframing cases, we validate the strengths of RAVA, showcasing its promise in the field of AI-powered video editing.
\end{itemize}

\section{Related Work}
\subsection{Video Editing}
Efforts in the field of movie analysis have made significant strides, especially in the realm of Audio-Visual Event (AVE) Localization. This particular task requires the identification and precise localization of various events within a video~\cite{tian2018audio,geng2023dense}. Such advancements are beneficial to video editors, as they can simplify the editing workflow~\cite{serrano2017movie}. It's important to note, however, that these studies do not provide a means for direct video editing. In addition to this line of research, some scholars have directly incorporated machine learning techniques into video editing. Argaw et al. introduce a benchmark suite specifically designed for video editing tasks, which includes but is not limited to, visual effects. This suite also facilitates the automatic organization of footage and provides assistance in video assembly~\cite{argaw2022anatomy}. Furthermore, Rao et al. present another benchmark aimed at selecting the best camera angle from multiple options, a crucial element in the production of television shows~\cite{rao2022temporal}. Despite these advancements, current methods do not address the challenge of video reframing, which involves highlighting and focusing on the most compelling segments of a video. The task of Video Salient Object Detection could potentially resolve this issue~\cite{hu2023tinyhd,yuan2023isomer,su2023unified}. However, the effectiveness of these methods is hampered by their reliance on specific training datasets, which limits their generalizability in diverse real-world settings and affects their interpretability.

\subsection{Open Vocabulary Segmentation}
Open Vocabulary Segmentation aims to segment images into meaningful regions without being constrained by a predefined set of categories. This approach significantly diverges from traditional segmentation methods~\cite{pinheiro2015image,lambert2020mseg,xian2019semantic}, which rely on a fixed vocabulary of labels, thus limiting their ability to generalize to novel or unseen objects. Seminal works like CLIP~\cite{radford2021learning} and ALIGN~\cite{jia2021scaling} enable segmentation models to identify and categorize a diverse array of unseen objects by leveraging natural language descriptions. Building on these foundations, LSeg~\cite{li2022languagedriven} trains an image encoder to create pixel embeddings and uses CLIP~\cite{radford2021learning} text embeddings as the per-pixel classifier. To make use of cheap image-level supervision, OpenSeg~\cite{ghiasi2022scaling}  employs weakly-supervised grounding loss and random word dropout to foster alignment between words and image regions. Although significant progress has been made, this field still faces numerous challenges and a scarcity of training data to address them. To cope with it, SAM~\cite{kirillov2023segment} introduces a pre-trained promptable foundation model for image segmentation, showcasing remarkable improvements in segmentation performance and adaptability. Recently, HQ-SAM~\cite{ke2024segment} demonstrates an architecture that closely integrates and utilizes the existing knowledge within the SAM structure. This approach enables the production of higher-quality masks while maintaining zero-shot capabilities. Studies like MedSAM~\cite{ma2024segment} also highlight the significant potential of SAM in the field of medicine.

\subsection{LLM Agent}
In recent times, we've seen the emergence of valuable frameworks such as AutoGPT~\cite{yang2023auto}, MetaGPT~\cite{hong2023metagpt}, and HuggingGPT~\cite{shen2024hugginggpt}, which symbolize the trend towards the swift integration of Large Language Models (LLMs) for performing intricate tasks. The development of multimodal LLMs, as referenced in works like Flamingo~\cite{alayrac2022flamingo}, Multimodal~\cite{furuta2023multimodal}, and AudioLM~\cite{shu2023audio}, expand the application spectrum of LLMs by enabling them to process diverse inputs such as text, images, audio, and video. This advancement allows models to directly handle multimodal inputs, moving beyond systems like TaskMatrix~\cite{liang2023taskmatrix}, which depend on several base models to convert visual information into linguistic forms through image captioning or object recognition. Capitalizing on the sophisticated perceptual abilities of these models, innovative approaches such as AppAgent~\cite{yang2023appagent}, MobileAgent~\cite{wang2024mobile}, and VisualWebArena~\cite{koh2024visualwebarena} are designed to precisely interact with mobile applications and execute web-based tasks. While there is a surge in LLM agent research, the domain of video editing is relatively untapped. LAVE~\cite{wang2024lave} serves as an agent for video editing, but its functions are limited to following user-defined goals. Our proposed research, however, aims to delve deeper into the potential of LLMs to enhance automated video reframing capabilities.

\section{Reframe Any Video Agent}

\begin{figure}[t]
\centering
\includegraphics[width=1\textwidth]{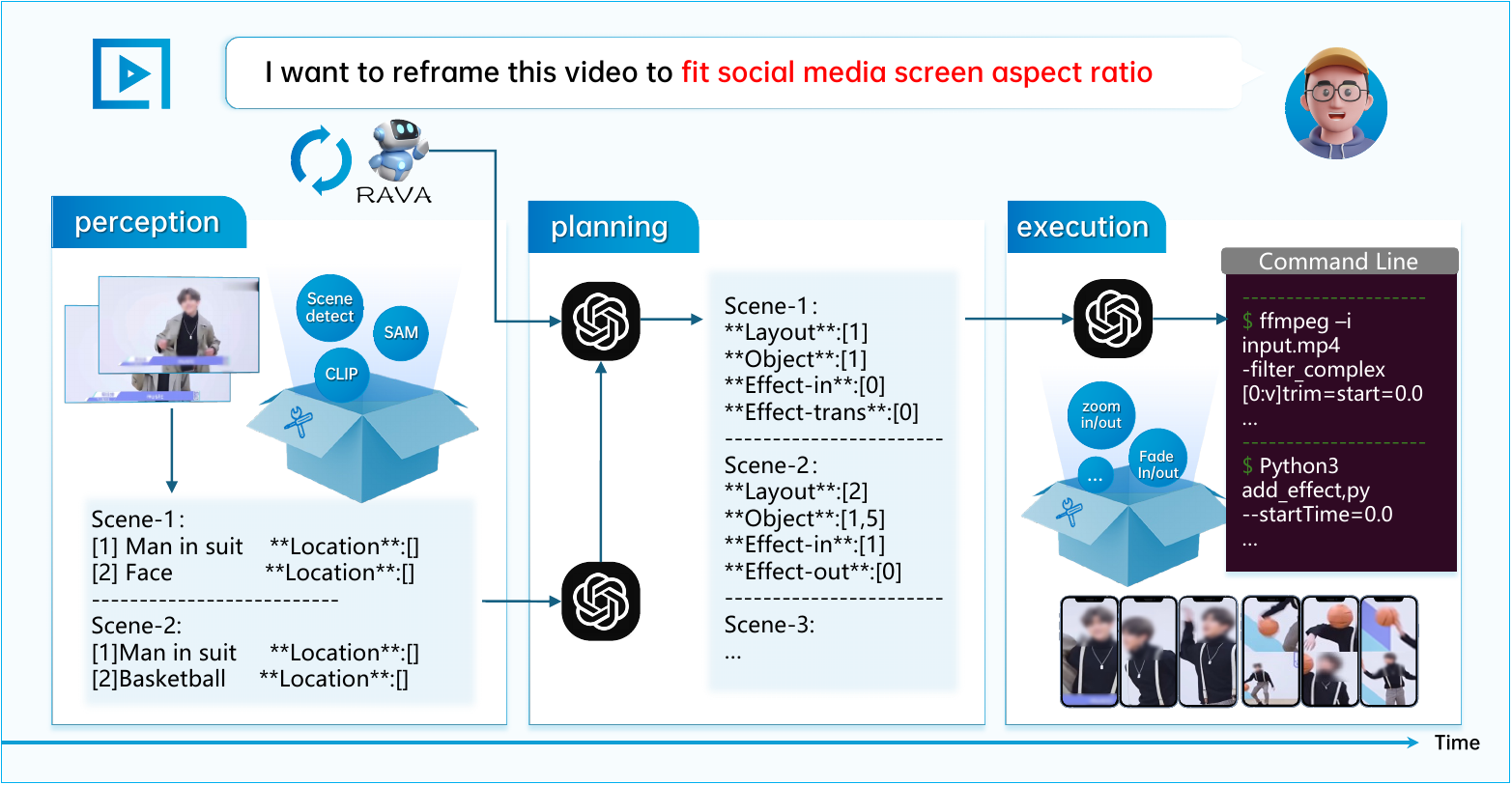}
\caption{The overall workflow of our proposed \textbf{R}eframe \textbf{A}ny \textbf{V}ideo \textbf{A}gent (RAVA). RAVA is capable of receiving user dialogue inputs with a language user interface (LUI) tailored for reframing tasks, and invokes the grounding function to retrieve object information within the video, then reframes the video automatically following the request from the users. }
\label{fig:method}
\end{figure} 

We present \textbf{R}eframe \textbf{A}ny \textbf{V}ideo \textbf{A}gent (RAVA) for the task of re-framing real-world videos, which also incorporates an LLM-based method supporting Language User Interface (LUI). Our method undergoes testing in an open-world setting, where scenes in videos might feature unseen objects. It is designed to robustly identify every object within a scene and discern which is of importance. Subsequently, it reframes the original video frames into varying aspect ratios, adapting to the specifications demanded by different social media applications or platforms. Furthermore, our method supports visual effects in two distinct scenarios: within individual scenes (in-scene) and between consecutive scenes (trans-scene). The entire workflow for the video reframing task can be accomplished automatically, specifically encompassing the following three key steps:

{\bf Object Grounding.} Specifically, for a given original video, there will exist $M$ scenes, and each video can be represented by $\{\mathbf{S}_{1}, \ldots, \mathbf{S}_M\}$, each scene composed of $N$ visual elements, formatted as $\{\mathbf{O}_{1}, \ldots, \mathbf{O}_N\}$, where $\mathbf{O}_{i}$ denote the segmentation mask of ${i}_{th}$ element. Our task is to identify the most significant object(s) like $\{\mathbf{O}_{i}, \ldots, \mathbf{O}_{j}\} \in \{\mathbf{O}_{1}, \ldots, \mathbf{O}_N\}$  within ${k}_{th}$ scene $\mathbf{S}_{k}$. 

{\bf Layout Setting.} In scenarios where multiple important objects are present, such as in a dialog scene, it is also necessary to determine the layout $\mathbf{L}_{k}\in\{1,2,3, \ldots, N\}$ - that is, how to simultaneously display multiple targets by sub-windows. Ultimately, for each scene, we aim to achieve a setting of $\mathbf{S}_{k}$ is $\mathbf{L}_{k}=n$, with the significant objects $\{\mathbf{O}_{i}, \ldots, \mathbf{O}_{j}\}$, where $n=count\{i,\ldots,j\}$.

{\bf Effect Adding.} Following this, our approach involves adding specific visual effects based on the content of the video. We introduce two types of visual effects: one that is applied within the current scene $\mathbf{S}_{k}$, such as zooming in and out, and another that is employed during scene transitions from $\mathbf{S}_{k}$ to $\mathbf{S}_{k+1}$, such as fading in and out.

Figure \ref{fig:method} demonstrates RAVA's workflow, three phases automate the task of reframing the video, adapting it to various aspect ratios required by different social media platforms or specifications, and enhancing viewer engagement through strategic layout decisions and visual effects.

\subsection{Perception}
For the perception phase of the agent, it can be divided into two segments: language learning and video understanding. Language learning focuses on comprehending the focal points of user interest, while video understanding centers on deciphering the content within the video frames.

Dialogue is an aspect where LLMs excel. By configuring prompts, the agent can comprehend the search objectives of the user. This process can be viewed as a structuring of LUI input information.  Specifically, we initiate the process by providing RAVA with a video and user interest. This context can include both the human-generated text query and the information retrieved from tools, as detailed subsequently. We also enable the LLM to output the video topic along with structured targets, which are then incorporated into the planning phase.

Inspired by films and scripts, we have adopted the shot detection approach to understand the entire video at the scene level. The specific methodology involves the use of scenedetect\footnote{https://www.scenedetect.com/} to transform the original video into a combination of multiple scenes $\{\mathbf{S}_{1}, \ldots, \mathbf{S}_M\}$. 

Understanding the video scenes necessitates the use of tools. Firstly, the RAM\cite{zhang2023recognize} identifies all objects within the scenes. Subsequently, we employ SAM~\cite{kirillov2023segment} and Grounded-SAM~\cite{ren2024grounded} to extract the masks of all objects and locations. CLIP\cite{radford2021learning} is utilized to acquire captions for each object. $\{\mathbf{O}_{1}, \ldots, \mathbf{O}_N\}$, where $\mathbf{O}_{i}$ denote the ${i}_{th}$ object, which is represented by the caption, mask, and location $\{{x}_{1},{y}_{1},{x}_{2},{y}_{2} \}$.  

Ultimately, the visual semantics of each video scene are transformed into a textual description. For instance, as demonstrated in Figure \ref{fig:method}, a scene in the video is described as "Scene-1:Object-1: a boy standing in...". This methodology allows for a comprehensive understanding and reframing of video content through an LLM agent by combining language comprehension and video analysis techniques.

\subsection{Planning}
Following perception, which provides a textual description and comprehension of the video content, the planning phase is pivotal as it involves devising a comprehensive strategy for reframing the video. This strategy should accommodate different aspect ratios, highlight imperative objects, and integrate visual effects in a cohesive manner that enhances user engagement. 

In this section, we detail the algorithms and methodologies employed for planning in the Reframe Any Video Agent (RAVA). With the insights gained from the perception phase, the planning phase consists of the following components:

{\bf Aspect Ratio Determination.} A critical part of planning is determining the desired aspect ratios for the output videos. We consider user preferences, platform requirements, and the context of the scenes. A dynamic decision-making process chooses an optimal aspect ratio for each scene to ensure the visual content is delivered effectively.

{\bf Object Importance Hierarchy.} The multiple objects identified in the perception phase need to be prioritized based on their significance relative to the context of the scene and user interest. Employing the language comprehension abilities of the LLM, we construct an importance hierarchy to aid in selection and layout decisions.

\begin{figure}[t]
\centering
\includegraphics[width=1\textwidth]{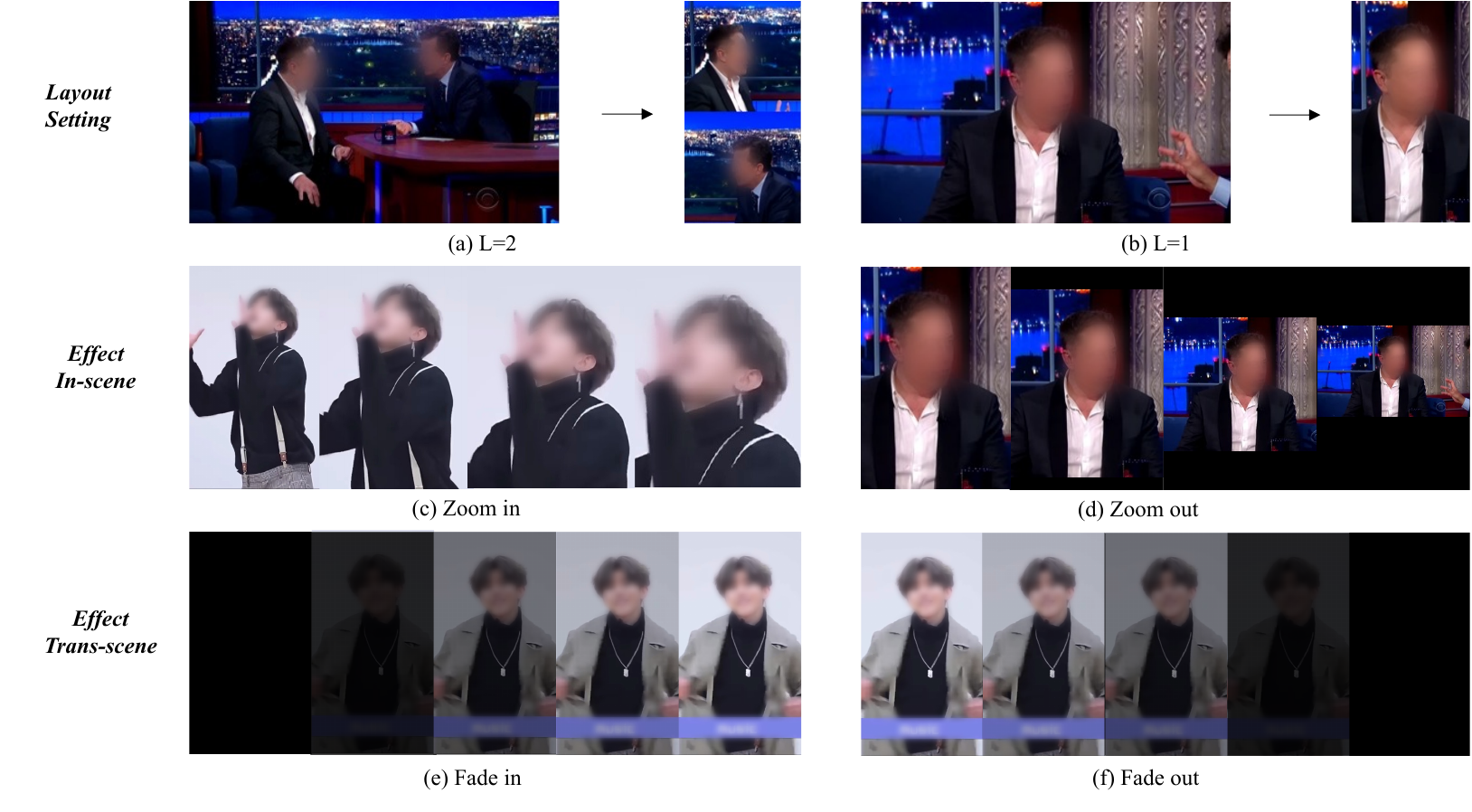}
\caption{The video editing tools in RAVA are described as follows: The first line is about `Layout settings', where `L' determines the number of selected objects in the video. The second line is `Effect In-scene', represents the visual effects within a scene, divided into `Zoom in' and `Zoom out'. The third line is `Effect Trans-scene', indicates the visual effects for scene transitions, divided into `Fade in' and `Fade out'.}
\label{fig:reframe workflow}
\end{figure}

{\bf Dynamic Layout Configuration.} Based on the significance and spatial arrangement of objects in a scene, we must plan for a layout that maximizes visual appeal and narrative coherence. As demonstrated in Figure \ref{fig:reframe workflow}, dynamic layout configurations consider dialogue exchanges, object interactions, and scene transitions to determine how the objects should be framed.

{\bf Visual Effect Strategy.} With visual effect preferences acquired from the perception phase, a plan is formulated to apply in-scene and trans-scene effects in a way that complements the narrative flow. The challenge lies not only in deciding which effects to use but also in determining their intensity and timing to maximize impact without distracting from the core content.

{\bf Execution Blueprint.} The culmination of the planning phase is an execution blueprint, which is a structured set of instructions ready to be parsed for execution. The blueprint encapsulates aspect ratios, object arrangement, effect schematics, and other relevant parameters.

{\bf Agent Feedback Loop.} An optional feedback loop allows the LLM to refine the planning based on a review of preliminary reframing results. This review process involves generating a low-resolution quick preview of the reframe and running it through the LLM for appraisal against the user objectives.

The planning phase incorporates algorithms for scene understanding to create a storyboard-like sequence of actions. This storyboard guides the execution phase, ensuring a seamless transition from a conceptual model to the practical implementation of video reframing. Each scene is treated as a separate module, with transitions planned to ensure a cohesive final product that aligns with the intentions of user and maximizes viewer engagement. RAVA's planning phase synthesizes all available data into a coherent plan, readying the system for precise execution.

\subsection{Execution}

The agent then proceeds to carry out the reframe task according to the plan generated before. During the execution phase, regular expression matching is employed to extract structured execution steps from the plan. These structured texts correspond to specific executable functions. We generate a JSON file for each video represented as $\{\mathbf{S}_{1}, \ldots, \mathbf{S}_M\}$ by scene detection, for the ${k}_{th}$ scene $\mathbf{S}_{k}$,  there are settings as layout $\mathbf{L}_{k}\in\{1,2,3, \ldots, N\}$, objects set $\{\mathbf{O}_{i}, \ldots, \mathbf{O}_{j}\}$, where $\mathbf{L}_{k}=count\{i,\ldots,j\}$, $\mathbf{E}_{in}\in\{zoom\ in, zoom\ out\}$, $\mathbf{E}_{trans}\in\{fade\ in, fade\ out\}$, This paper presents several simple implementations of special effects and argues that our framework is extensible.

For instance, if the layout is set to 2, it will choose the two objects as the main subjects of the video and arrange them vertically. If the in-scene visual effects $\mathbf{E}_{in}$ column has selected $\{zoom\ in\}$, the agent will invoke the corresponding API function to magnify the target(s). Additionally, if the trans-scene $\mathbf{E}_{trans}$ visual effects is $\{fade\ out\}$, a transition effect will also be inserted at the end of the current scene.

\section{Experiments}
To validate the effectiveness of the Reframe Any Video Agent (RAVA) we proposed, we conduct evaluations through two principal tasks. In the first task, we employ the model to address a significant challenge in the field of video understanding, namely video salient object detection. This involves segmenting the most visually prominent objects as perceived by human vision. In the second task, we apply the model to tackle the task of video reframing. This process involves adjusting the video frame to focus on the most important elements, enhancing the overall composition and storytelling aspect of the visual content.

\begin{table}[t]
\centering
\caption{Our experimental results for Video Salient Object Detection task. The abbreviation `SD' denotes `Scene Detection'. }
\label{tab:VSOD-results}
\begin{tabular}{@{}ccccccccccc@{}}
\toprule
\multicolumn{1}{c}{\multirow{2}{*}{Method}} & \multicolumn{2}{c}{SD}                    & \multicolumn{4}{c}{DAVIS$_{16}$}         & \multicolumn{4}{c}{FBMS}          \\ \cmidrule(lr){2-3} \cmidrule(lr){4-7} \cmidrule(lr){8-11}
\multicolumn{1}{c}{}                        & $\alpha_1$              & $\alpha_2$              & ~MAE~    & max-$F_\beta$ & max-$E_m$ & $S_m$     & ~MAE~    & max-$F_\beta$ & max-$E_m$ & $S_m$     \\ \midrule
UPL                                  & \multirow{3}{*}{5}  & \multirow{3}{*}{5}  & .0390  & .8025 & .9183 & .8426 & .0850  & .6651 & .8513 & .7439 \\
A2S-v2                                      &                     &                     & .0663 & .4858 & .5786 & .5817 & .0851 & .6444 & .8366 & .7004 \\
Ours                                        &                     &                     &  .0501      &  .7025  & .8219      & .7795   &  .1015  & .5721  & .7532 & .6643 \\ \cmidrule{1-11}
UPL                                  & \multirow{3}{*}{5}  & \multirow{3}{*}{30} & .0367 & .8127 & .9275 & .8481 & .0844 & .6673 & .8458 & .7527 \\
A2S-v2                                      &                     &                     & .0638 & .5046 & .5929 & .5926 & .0832 & .6406 & .8288 & .7054 \\
Ours                                        &                     &                     &  .0419      &  .6727   &   .8177     &  .7680 & .1148  & .5446  & .7131 & .6422  \\ \cmidrule{1-11}
UPL                                  & \multirow{3}{*}{10} & \multirow{3}{*}{5}  & .0381 & .8009 & .9210  & .8361 & .0848 & .6670  & .8615 & .7373 \\
A2S-v2                                      &                     &                     & .0640  & .4907 & .5723 & .5836 & .0900   & .6299 & .8446 & .6836 \\
Ours                                        &                     &                     &  .0506  &  .7126  &  .8256 &  .7804 & .1313  & .5128 & .6776 &.6089\\ \bottomrule
\end{tabular}
\end{table}

\subsection{Video Salient Object Detection}
\subsubsection{Datasets.}
Our approach is scrutinized through the lens of two widely used datasets, DAVIS$_{16}$ \cite{perazzi2016benchmark} and FBMS \cite{ochs2013segmentation}. The former consists of 50 videos, amounting to 3,455 annotated frames. The latter dataset incorporates 33 supplementary video sequences, collectively characterized by 720 annotated frames.
\subsubsection{Metrics.}
\vspace{-10pt}
Four widely used evaluation metrics are employed to assess the performance, including Mean Absolute Error (MAE)~\cite{perazzi2012saliency}, F-measure ($F_\beta$)~\cite{achanta2009frequency}, E-measure ($E_m$)~\cite{fan2018enhanced}, and S-measure ($S_m$)~\cite{fan2017structure}.
\subsubsection{Settings.} 
\vspace{-10pt}
We composite the frames into videos at 30 frames per second. Then, The scene detection method is applied for video splitting. To minimize the possibility of a scene detection result of 0 and to reduce the number of scenes detected under each category, we choose the parameter combination of a lower threshold called $\alpha_{1}$ and a higher detection interval called $\alpha_{2}$, corresponding to parameters named threshold and min scene length. Following this phase, each scene is individually subjected to respective testing procedures to generate the salient masks. It is worth noting that each object $\mathbf{O}_{i}$ in the perception phase generates only a caption and a mask. This is because over-adjusting the frame can lead to an undesirable jittery effect, which can negatively impact the overall viewing experience.
\subsubsection{Results.}
\vspace{-10pt}
\begin{figure}[t]
    \centering
    \includegraphics[width=1\linewidth]{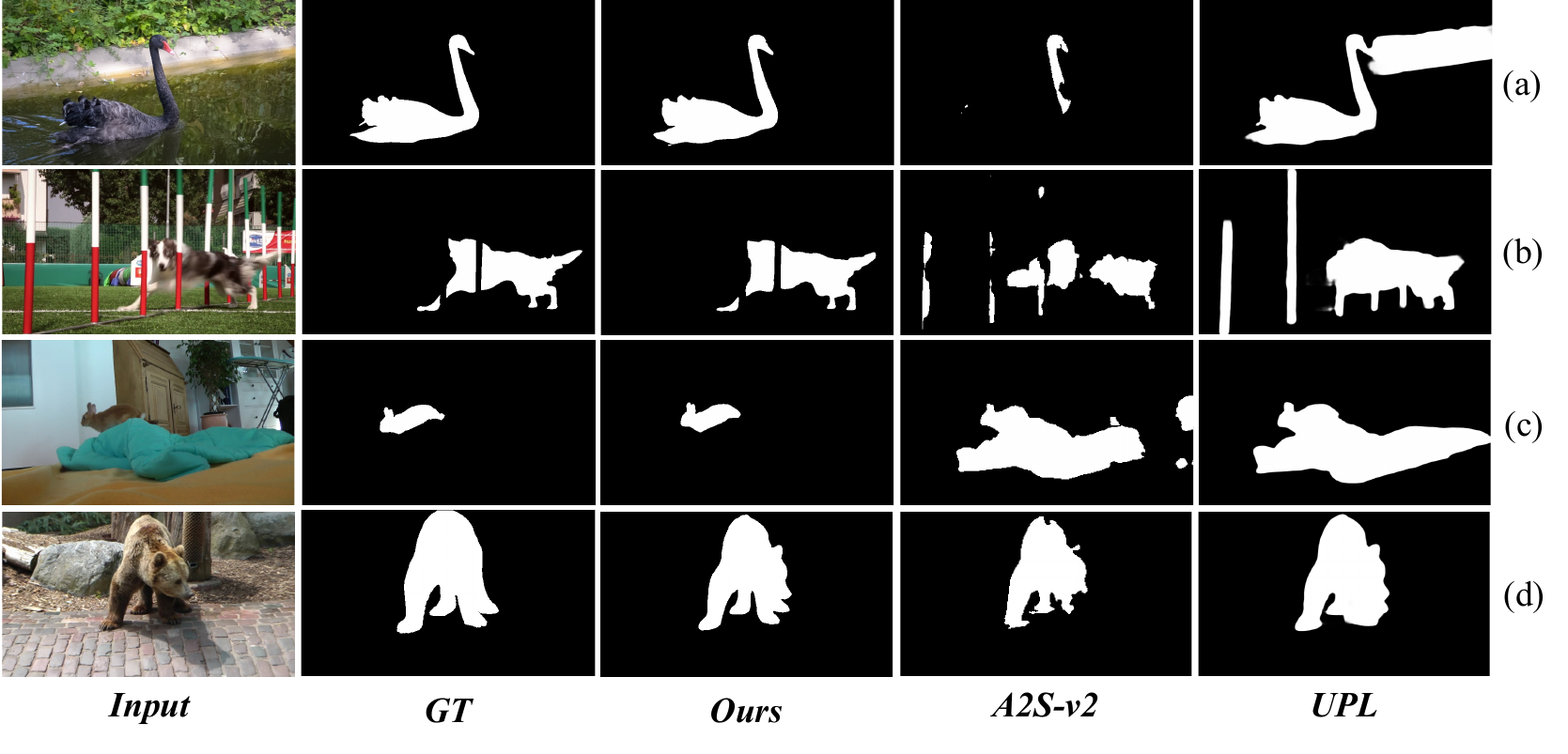}
    \caption{The qualitative results on two video salient object detection datasets in comparison with two state-of-the-art methods. The findings indicate that RAVA copes well even in the presence of occlusions and distractions, and at times, it even surpasses the results of human annotations.}
    \label{fig:vsod}
    \vspace{-5pt}
\end{figure}

We compare RAVA with the current state-of-the-art video salient object detection methods, namely UPL~\cite{yan2022unsupervised} and A2S-v2~\cite{zhou2023texture}. As can be seen in Table~\ref{tab:VSOD-results}, under various settings of scene detection parameters, our approach consistently achieves competitive outcomes. It is noteworthy that our framework is not specifically designed for the task of video salient object detection. The fact that it achieves competitive results is sufficient to demonstrate the effectiveness of RAVA. To further analyze our results, we present a subset of visualized results in Figure \ref{fig:vsod}: 

(a) RAVA is capable of precisely segmenting the entirety of the Blackswan instance; however, A2S-v2 fails to provide a complete mask, and UPL erroneously segments a part of the background as if it were part of the Blackswan. 

(b) RAVA maintains accurate segmentation even when objects are occluded, whereas A2S-v2, despite attempting to address occlusions, incorrectly segments parts of the occluding item; UPL, on the other hand, overlooks the occlusion altogether.

(c) When confronted with the distraction of other objects in the scene, both A2S-v2 and UPL fail to respond accurately, while RAVA can precisely identify the salient object.

(d) This example better demonstrates the robust capability of RAVA, which, through a powerful segmentation model, can perceive and achieve results that are more accurate than human annotations.

\subsubsection{Bad Cases.}
\begin{figure}[t]
    \centering
    \includegraphics[width=1\linewidth]{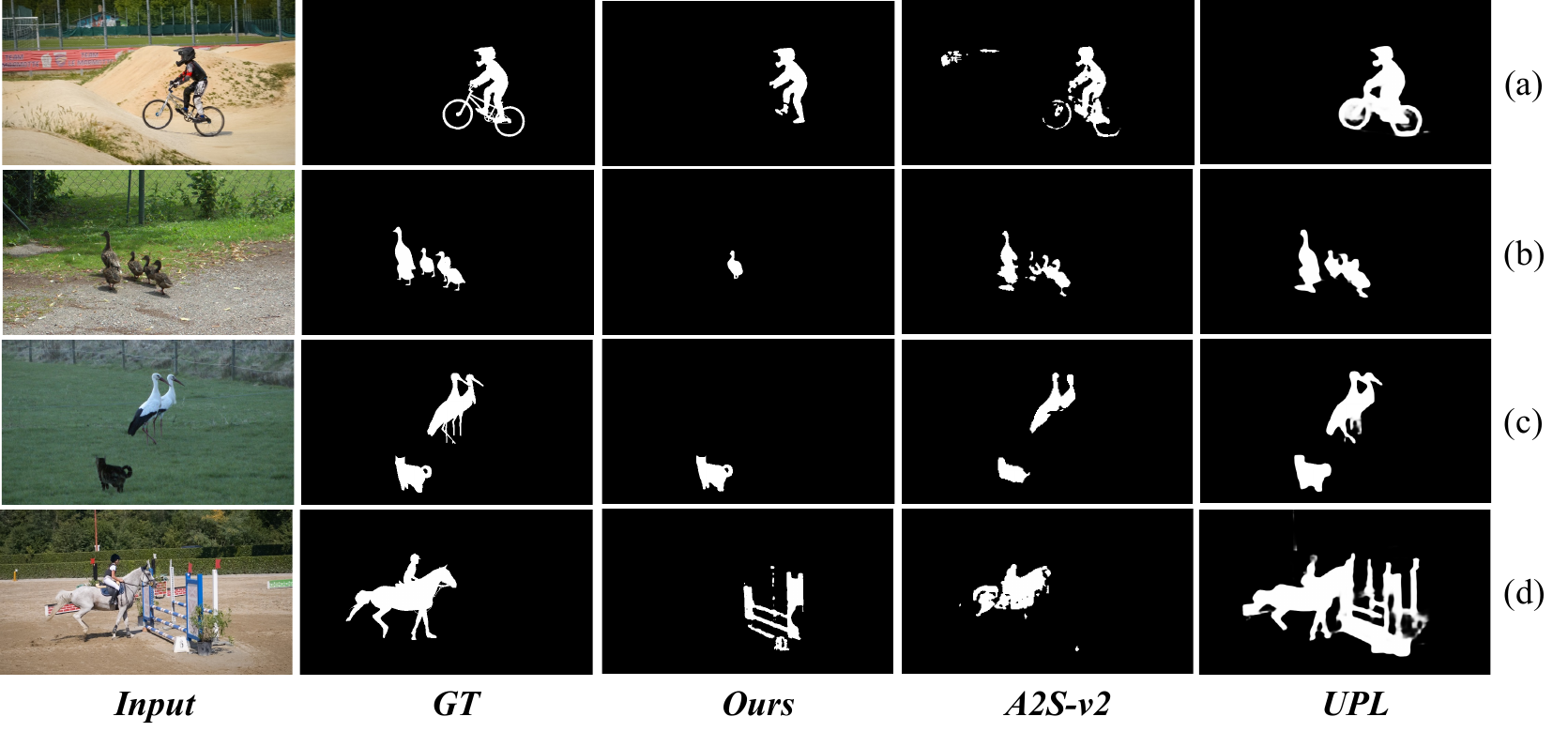}
    \caption{The cases where using RAVA might fail include situations (a) and (b), where it struggles to effectively handle composite objects and clusters of closely situated objects. However, in cases (c) and (d), despite the differences in perception of salient objects, we consider the segmentation results to be acceptable.}
    \label{fig:bad_case}
    \vspace{-10pt}
\end{figure}

Even though RAVA exhibits commendable performance across most scenarios, there are still some bad cases. We believe these instances contribute to its inability to reach state-of-the-art status in the task of video salient object detection. These results are illustrated in Figure \ref{fig:bad_case}:

(a) In this case, RAVA can segment the person riding the bicycle with relative precision, but it fails to recognize that the bicycle and the person constitute a single instance, therefore not achieving an optimal result.

(b) When multiple objects close to each other that should be considered as one instance, RAVA struggles to include the other objects within the segmentation.

(c) RAVA segments the black cat in the image as the salient object, while ignoring the two birds that are farther away. We believe this result is reasonable, as perspectives on what constitutes a salient object can vary. This process still demonstrates the visual understanding of RAVA.

(d) Although the GT provided in this image depicts the person and the horse as a single entity, the fences are also prominently marked in blue, making it reasonable to consider the fences as the salient objects. This scenario is similarly observed in the mask produced by the UPL, which segments out both the fences and the horse.

\subsubsection{Ablation Study.}
To further evaluate the effectiveness of RAVA in a single-language modality, we substituted the LLM with GPT-4~\cite{openai2023gpt4}, which lacks multimodal perception capabilities. Apart from omitting visual inputs in the perception, we maintain all other settings unchanged.

Table.\ref{tab:VSOD-gpt4} reveals that even in the absence of direct visual information, GPT-4 is still capable of delivering commendable performance on two datasets based on the textual descriptions provided. This underscores the robust transferability of RAVA. However, considering the existing gap, it is necessary to use a LLM with strong visual perception capabilities.
\begin{table}[t]
\centering
\caption{Our experimental results for Video Salient Object Detection task on single-modality LLM. The abbreviation `SD' denotes `Scene Detection'. }
\label{tab:VSOD-gpt4}
\begin{tabular}{@{}ccccccccccc@{}}
\toprule
\multicolumn{1}{c}{\multirow{2}{*}{LLM}} & \multicolumn{2}{c}{SD}                    & \multicolumn{4}{c}{DAVIS$_{16}$}         & \multicolumn{4}{c}{FBMS}          \\ \cmidrule(lr){2-3} \cmidrule(lr){4-7} \cmidrule(lr){8-11}
\multicolumn{1}{c}{}                        & $\alpha_1$              & $\alpha_2$              & ~MAE~    & max-$F_\beta$ & max-$E_m$ & $S_m$     & ~MAE~    & max-$F_\beta$ & max-$E_m$ & $S_m$     \\ \midrule
\multirow{3}{*}{GPT-4}                                & 5  & 5  & .0831  & .6497 & .7885 & .7395 & .1294  & .4953 & .6943 & .6125 \\
                                      &    5        &   30      & .0548 & .6163 & .7930 & .7422 & .1642 & .4856 & .6670 & .5915 \\
                                        &      10   &     5     &  .0690      &  .6432  & .7913      & .7454   &  .1307  & .4931  & .6684 & .6069 \\
\bottomrule
\end{tabular}
\end{table}

\subsection{Video Reframing}
\subsubsection{Settings.}
To assess the video editing capabilities of RAVA in the wild, a user study is conducted with 12 participants. Edited versions of 5 videos are created using three reframe methodologies in addition to RAVA: 
\begin{itemize}
\item \textbf{Editor}: This method involves a professional video editor (experience >3 years) who manually reframe the videos. 
\item \textbf{Adobe}: This method is based on the results obtained by ordinary users utilizing the reframe tool in Adobe Premiere Pro to adjust the videos, following the instructions\footnote{https://helpx.adobe.com/premiere-pro/using/auto-reframe.html}.
\item \textbf{Center Cut}: This method selects the center point of the video, maintaining a 9:16 aspect ratio, with the width unchanged. 
\end{itemize}
To minimize the impact of the video itself and to maintain an element of unbiased evaluation by users, the open caption tool\footnote{https://www.opus.pro/tools/opusclip-captions} is employed to add captions for each video. After watching the original video, each participant views the 4 edited versions in a random sequence. The participants review all reframed videos, and they are unaware of the editing methods employed for each video. This arrangement led to a comprehensive experimental design involving 5 (number of videos) × 12 (users) × 4 (editing strategies). Users are required to compare the reframed version of a video with the original and provide a rating on a scale from 0 to 5 for each of the attributes. These attributes were inspired by studies on video re-positioning\cite{moorthy2020gazed}, and it’s important to note that, although video reframing and video repositioning differ technically, both aim to direct the attention of the viewer to the focal scene events within given rendering constraints. Thus, some of the questions used to assess methods of video re-positioning are also applicable to video editing. Our attributes of interest include: 
\begin{itemize} % Adjust left margin as needed
    \item \textbf{Content Preservation}
    \begin{itemize}
        \item (Relevance) How well does the reframed video maintain the key elements of the original content?
        \item (Completeness) Are important parts of the scene, especially the main subjects, preserved in the reframed version?
    \end{itemize}

    \item \textbf{Continuity and Consistency}
    \begin{itemize}
        \item (Continuity) Does the sequence of shots follow a logical temporal order without visually jarring jumps?
        \item (Consistency) Does the reframing maintain the original style and mood of the video? 
    \end{itemize}

    \item \textbf{User Experience}
    \begin{itemize}
        \item (Satisfaction) Overall, how satisfied are the users with the reframed video?
        \item (Usability) Overall, are you willing to post this video on your social platform?
    \end{itemize}

    \item \textbf{Technical Quality}
    \begin{itemize}
        \item (Resolution and Clarity) Is the video crisp and clear, without degradation from cropping or zooming?
        \item (Stability) Does the video maintain stability, without introducing shakiness due to reframing?
    \end{itemize}
\end{itemize}

\begin{figure}[t]
    \centering
        \begin{subfigure}[b]{0.24\linewidth}
        \includegraphics[width=\linewidth]{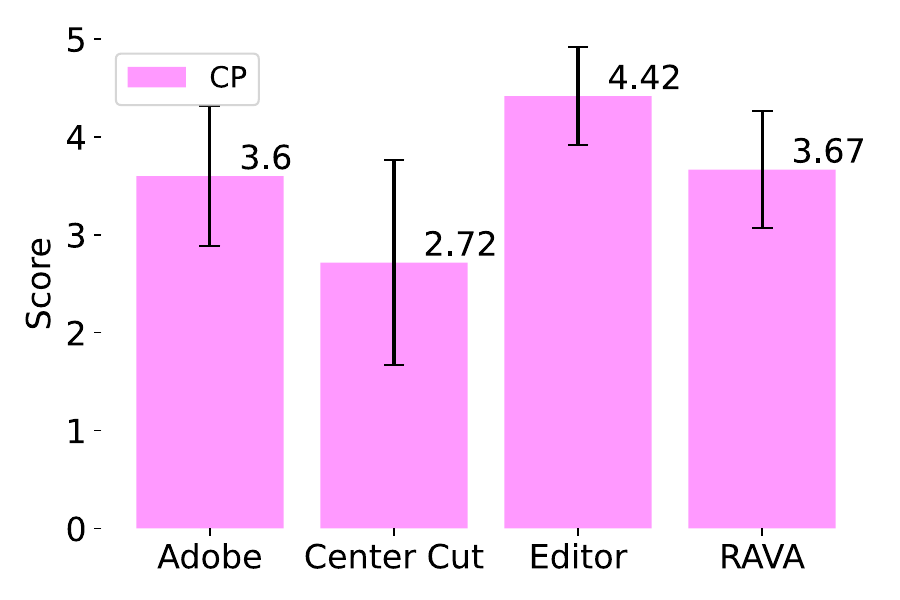}
        \caption{CP}
        \label{fig:CP}
    \end{subfigure}
    \hfill % 
    \begin{subfigure}[b]{0.24\linewidth}
        \includegraphics[width=\linewidth]{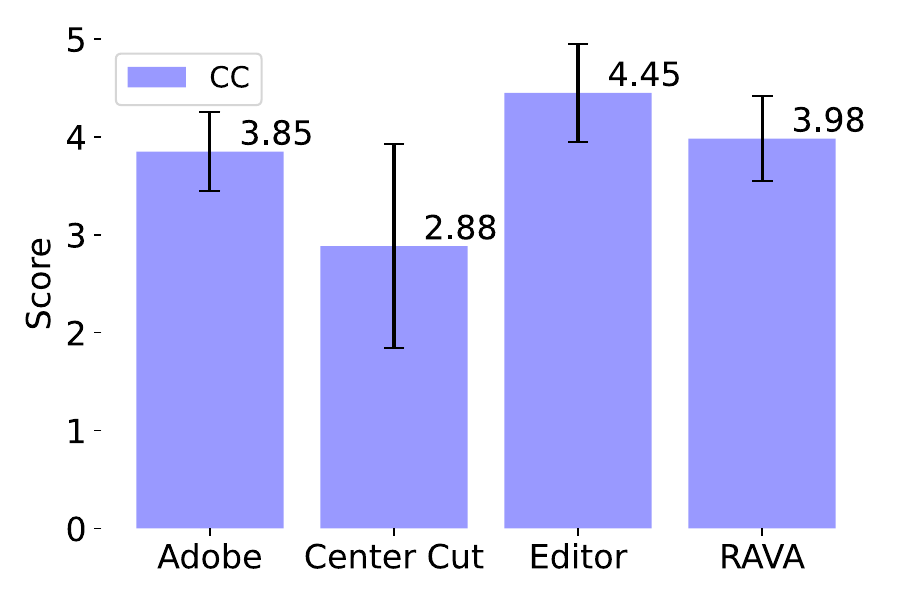}
        \caption{CC}
        \label{fig:CC}
    \end{subfigure}
    \hfill % 
    \begin{subfigure}[b]{0.24\linewidth}
        \includegraphics[width=\linewidth]{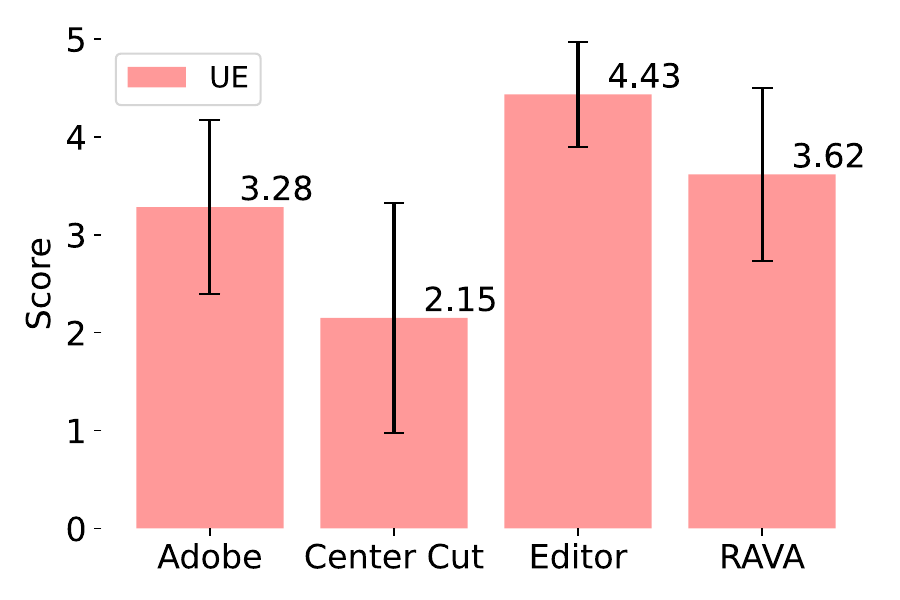}
        \caption{UE}
        \label{fig:UE}
    \end{subfigure}
    \hfill % 
    \begin{subfigure}[b]{0.24\linewidth}
        \includegraphics[width=\linewidth]{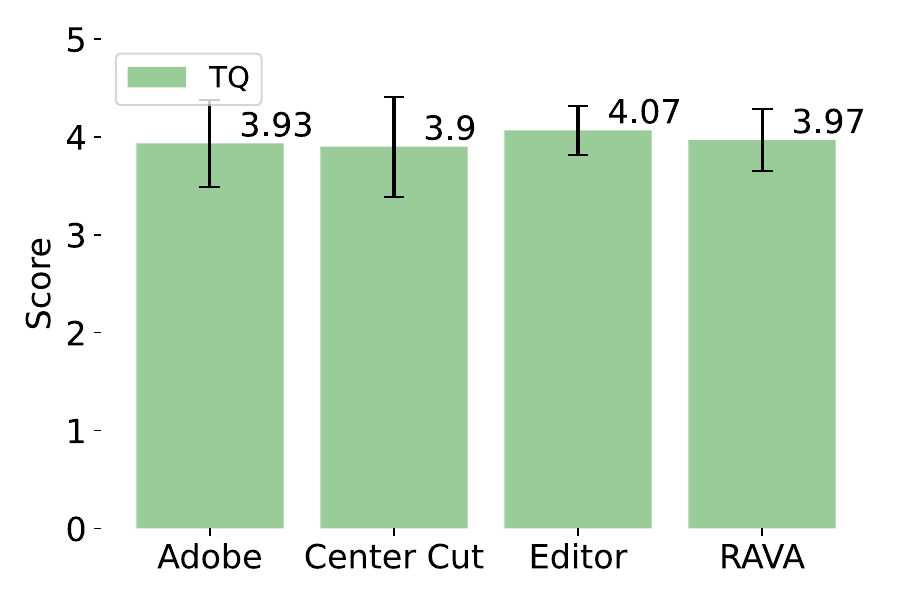}
        \caption{TQ}
        \label{fig:TQ}
    \end{subfigure}
    \caption{Overall scores of the individual attributes: Content Preservation (CP), Continuity and Consistency (CC), User Experience (UE), and Technical Quality (TQ).}
    \label{fig:concatenated_output}
\end{figure}
  
\subsubsection{Results.}
As seen in the results presented in Figure \ref{fig:concatenated_output} and outlined in our experimental data, the traditional editing method, referred to as `Editor', received the highest overall mean score of $4.34$, indicating a strong ability to maintain the relevance and completeness of the original content. This could be attributed to the manual effort and expertise that video editors bring to the reframing endeavor, ensuring that significant elements are not lost. Our proposed method, RAVA, achieves an overall score of $3.81$ from four aspects, suggesting that while RAVA performs reasonably well in terms of relevancy and scene completeness, there is room for improvement when compared to professional editing. The performance of RAVA surpasses the automated `Adobe' tool, with `Adobe' scoring an overall score of $3.67$. This close competition hints that RAVA is on par with other available semi-automated video editing tools in terms of content preservation. The `Center Cut' received the lowest mean score of $2.72$ in Content Preservation, reflecting its limited ability to identify and maintain critical video elements. 

The variability in scores, as indicated by the interquartile range in the boxplot, further substantiates the need for an intelligent and context-aware re-framing technique. Future work could explore enhancing the object identification and importance determination algorithms of RAVA to further close the gap between automated and professional video editing tools.

Besides, we strongly recommend that readers view the edited video included in the supplementary materials for a more intuitive understanding.

\section{Conclusion}
We present a groundbreaking approach in the realm of video reframing by introducing \textbf{R}eframe \textbf{A}ny \textbf{V}ideo \textbf{A}gent (RAVA), a sophisticated agent powered by large language models (LLMs) that excels in executing video reframing tasks based on human instructions. Through a meticulous three-stage process encompassing perception, planning, and execution, RAVA showcases its prowess in understanding user directives, dissecting scenes, prioritizing objects, configuring layouts, and applying visual effects—all while ensuring alignment with narrative and user preferences. Our experiments, spanning from traditional computer vision tasks to real-world video reframing scenarios, demonstrate the efficacy and promise of RAVA in the domain of AI-driven video editing. By validating the capabilities of RAVA through quantitative results and user studies, we establish its potential as a transformative tool for enhancing content creation and resonating with diverse audiences across various platforms.

\subsubsection{Limitations.}
\vspace{-10pt}
The main constraint of our study stems from its dependence on the underlying performance of foundation models. Utilizing a more sophisticated visual model may lead to enhanced results. Additionally, exploring how to extend the agent to perform editing across the video timeline, thereby achieving the condensation of longer content into shorter formats, represents a promising direction for development.

\subsubsection{Social Impact.}
\vspace{-10pt}
The research presented in this paper involves the utilization of publicly available video content sourced from YouTube. We have ensured that all video materials employed in our study adhere to the platform's terms of service regarding non-commercial, research-based usage. 

\bibliographystyle{splncs04}
\bibliography{egbib}
\end{document}